\documentclass{article}
\usepackage[preprint]{spconf,amsmath,epsfig}
\usepackage{multirow}
\usepackage{color}
\usepackage{booktabs}

\copyrightnotice{\copyright\ IEEE 2024}

\makeatletter
\def\blfootnote{\xdef\@thefnmark{}\@footnotetext}
\makeatother

\title{Event-to-Video Conversion for Overhead Object Detection}
\name{Darryl Hannan\textsuperscript{\rm 1,4}\textsuperscript{$*$},
Ragib Arnab\textsuperscript{\rm 1,2}\sthanks{Co-Authors},
Gavin Parpart\textsuperscript{\rm 1},
Garrett T. Kenyon\textsuperscript{\rm 3},
Edward Kim\textsuperscript{\rm 4},
and Yijing Watkins\textsuperscript{\rm 1}}

\address{Pacific Northwest National Laboratory, Seattle, WA, USA\textsuperscript{\rm 1}\\
University of Texas at Dallas, Richardson, TX, USA\textsuperscript{\rm 2} \\
Los Alamos National Laboratory, Los Alamos, NM, USA\textsuperscript{\rm 3} \\
Drexel University, Philadelphia, PA, USA\textsuperscript{\rm 4}\\
darryl.hannan@pnnl.gov}
\begin{document}
%
\maketitle
\begin{abstract}
Collecting overhead imagery using an event camera is desirable due to the energy efficiency of the image sensor compared to standard cameras. However, event cameras complicate downstream image processing, especially for complex tasks such as object detection. In this paper, we investigate the viability of event streams for overhead object detection. We demonstrate that across a number of standard modeling approaches, there is a significant gap in performance between dense event representations and corresponding RGB frames. We establish that this gap is, in part, due to a lack of overlap between the event representations and the pre-training data used to initialize the weights of the object detectors. Then, we apply event-to-video conversion models that convert event streams into gray-scale video to close this gap. We demonstrate that this approach results in a large performance increase, outperforming even event-specific object detection techniques on our overhead target task. These results suggest that better alignment between event representations and existing large pre-trained models may result in greater short-term performance gains compared to end-to-end event-specific architectural improvements.
\end{abstract}
\section{Introduction}
\blfootnote{Copyright 2024 IEEE. Published in the 2024 Southwest Symposium on Image Analysis and Interpretation (SSIAI) (SSIAI 2024), scheduled for 17-19 March 2024 in Santa Fe, New Mexico, USA. Personal use of this material is permitted. However, permission to reprint/republish this material for advertising or promotional purposes or for creating new collective works for resale or redistribution to servers or lists, or to reuse any copyrighted component of this work in other works, must be obtained from the IEEE. Contact: Manager, Copyrights and Permissions / IEEE Service Center / 445 Hoes Lane / P.O. Box 1331 / Piscataway, NJ 08855-1331, USA. Telephone: + Intl. 908-562-3966.}
\begin{figure}[t]
    \centering
    \includegraphics[width=\columnwidth]{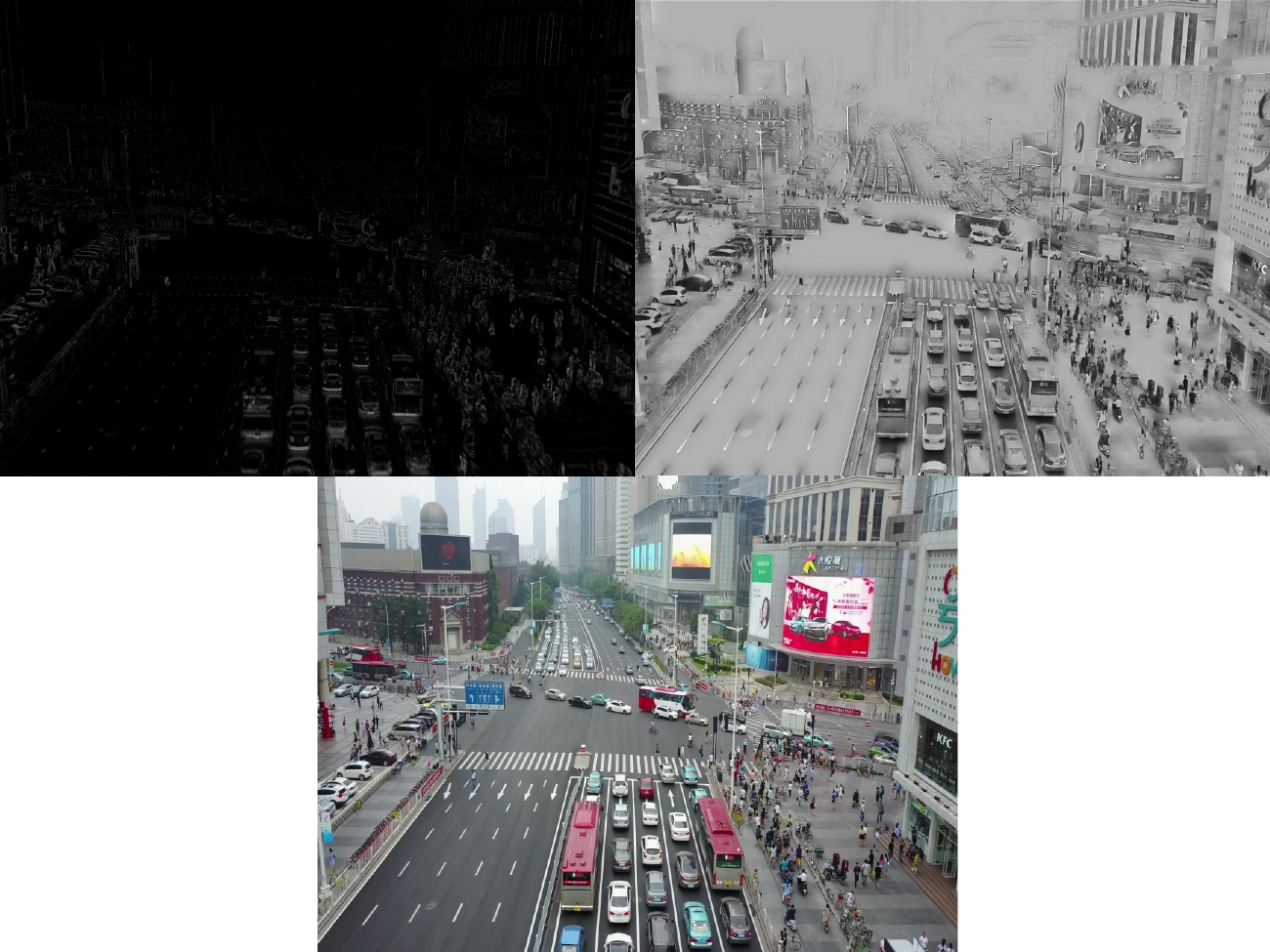}
    \caption{Comparison of the same VisDrone-VID \cite{visdrone} scene using various input representations. Top Left: Event Count Map. Top Right: FireNet \cite{firenet} Gray-scale Frame. Bottom: Original RGB Frame.}
    \label{fig:scene_comparison}
\end{figure}
Overhead object detection involves detecting objects in images where the camera is positioned above the objects of interest. This may include satellite imagery or images collected from an aerial vehicle. Detecting objects in an overhead setting is challenging for a number of reasons including the prevalence of small objects, a lack of publicly available, large-scale datasets, and a large domain shift compared to datasets commonly employed for pre-training object detectors. Furthermore, when considering model deployment in this setting, additional considerations need to be taken into account including the energy efficiency of the proposed system, the total weight of the device, whether to process the data on or off-device, and more.

One hardware component that can help overcome some of these practical constraints is an event camera. Event cameras are biologically-inspired image sensors that detect changes in brightness. Pixels in these sensors operate asynchronously, independently storing and capturing changes in brightness at a given pixel location over time. This results in a number of benefits including greater dynamic range, increased energy efficiency, and a much higher effective frame rate. However, these benefits come at a cost when considering the impact that the event camera has on downstream image processing. Event streams are extremely sparse compared to standard RGB frames, rendering standard image processing techniques ineffective.


In this paper, we investigate the utility of event cameras within the context of an overhead object detection task and seek to mitigate some of the issues associated with processing event streams using deep learning object detection algorithms. We present experiments comparing dense event representations to RGB frames using the VisDrone-VID dataset \cite{visdrone}, applying a variety of object detectors to both of these modalities, and illustrate that there is a large gap in performance between them. We then demonstrate through further experimentation that this gap is in part due to a large domain shift between standard vision datasets used for pre-training and the dense event representations. To overcome this problem, we propose leveraging FireNet \cite{firenet} and FlowNet \cite{stoffregen2020reducing} to convert the event streams into standard gray-scale videos. We demonstrate that this approach almost doubles performance. We compare our approach to object detection techniques specifically designed for event cameras and demonstrate that our more simplistic approach outperforms these models in our overhead setting. Our results suggest that future work should focus on improving the alignment between event stream representations and RGB frames to further take advantage of powerful, pre-trained deep learning models.

\section{Related Work}
A number of different approaches have been applied to event-based object detection.
The line of work more closely aligned with this paper involves applying deep learning techniques to dense event representations \cite{gen4,rvt}. Other works have explored deep learning-based object detectors in the context of event streams \cite{iacono2018towards,mechler2023transferring}. However, these do not focus on overhead video nor do they propose leveraging a conversion tool to better align event stream video with common pre-training datasets.
Event cameras have previously been applied to overhead video in other contexts. Much of the work in this space focuses on automatically piloting an aerial vehicle \cite{falanga2020dynamic,sanket2020evdodgenet,paredes2020back}. Our work distinguishes itself from these due to our explicit focus on object detection.

\section{Methodology} \label{sec:method}
The task that we consider is overhead object detection using an event stream. Consider an event stream $E$, where each event is a 4-tuple $(p, x, y, t)$, where $p$ is the polarity of the event (either positive or negative), $x$ is the horizontal pixel dimension, $y$ is the vertical pixel dimension, and $t$ is the time at which the event occurred. An agent receives $E$ as input and is tasked with producing a sequence of bounding boxes $B$ and a corresponding sequence of classes $C$, where each bounding box contains an object in the video at a given point in time along with the corresponding class. This task is more challenging than standard object detection due to the sparse nature of event streams. Many event cameras have microsecond resolution leading to long sequences where an overwhelming majority of the sequence will not contain any events. In scenes with limited movement, changes in brightness will be infrequent, potentially requiring the agent to combine information over long time spans.

For our initial experiments, we apply state-of-the-art object detection models to event count maps (ECMs). To create these maps we bin events into fixed time bins. Let $T$ be the total temporal length of $E$ and let our time-window be $w$. We segment $E$ into $n = T/w$ temporal bins, resulting in a sequence of bins $B = b_0, b_1, ..., b_n$. We then compute each ECM $m_n$ by summing over the temporal dimension within $b_n$. $m$ can then be interpreted as a single channel image, where each pixel location stores the average change in brightness over the time window. By further scaling these values to lie between 0 and 255, $m$ can be processed as a gray-scale image (see Figure \ref{fig:scene_comparison} Top Right).

In Section \ref{sec:experiments}, we demonstrate that ECMs are significantly inferior to RGB frames. Therefore, we propose leveraging an existing event-to-video conversion tool to improve the quality of the dense event representations. The event-to-video conversion models that we consider are FireNet \cite{firenet} and FlowNet \cite{stoffregen2020reducing}. FireNet is a convolutional recurrent architecture that operates over sequences of dense event representations. There is no striding or max pooling, resulting in the spatial dimensions being preserved throughout the model. Whereas, FlowNet is a U-Net architecture with a ConvLSTM layer immediately following the encoder. The presence of recurrent units allows both models to utilize information over long timescales. We present experiments using both the pre-trained weights provided by the original authors and our own fine-tuned weights produced by initializing the model with the pre-trained weights then doing additional training on our target overhead event dataset.

\section{Experiments and Results} \label{sec:experiments}
\subsection{Experimental Setup}
Collecting overhead event data with aligned RGB frames at scale is challenging. Therefore, we rely upon an existing overhead object detection dataset, the VisDrone-VID dataset \cite{visdrone}, for all of our experiments.
A sample frame can be seen in Figure \ref{fig:scene_comparison} (Bottom). To convert the VisDrone dataset into events, we use the v2e \cite{v2e} toolkit.

For building our object detectors we leveraged MMDetection \cite{mmdetection} and MMYOLO \cite{mmyolo}. We present results for 3 detectors: Cascade RCNN \cite{cascade}, YOLOv8 \cite{yolov8}, and DINO \cite{dino}. These models collectively represent a wide breadth of techniques, ensuring that our results are not biased towards a specific type of detector. Additionally, we explore an event-specific object detection approach, Recurrent Vision Transformer (RVT) \cite{rvt}, that achieves strong results on the Gen4 event-based object detection dataset \cite{gen4}. We use mean average precision (mAP) and AP 50 as our evaluation metrics.

\subsection{Initial Event Stream vs. RGB Frame Comparison}
\begin{table}[]
    \centering
    \begin{tabular}{l l l l l}
    \toprule
        Model & Input & mAP & AP 50 \\\midrule
        \multirow{2}{*}{DINO} & RGB Frame & 0.174 & 0.351 \\
        & ECM & 0.071 & 0.155 \\\hline
        \multirow{2}{*}{YOLOv8} & RGB Frame & 0.139 & 0.268 \\
        & ECM & 0.073 & 0.159 \\\hline
        \multirow{2}{*}{Cascade RCNN} & RGB Frame & 0.128 & 0.246 \\
        & ECM & 0.063 & 0.132 \\\hline
        RVT & ECM & 0.046 & 0.104 \\
        \bottomrule
    \end{tabular}
    \caption{VidDrone-VID dev test set performance for both original RGB frames and converted event count maps.}
    \label{tab:main_results}
\end{table}
We began by directly training and evaluating our object detection models on both the ECMs and RGB frames. The results can be found in Table \ref{tab:main_results}. We found that there was a large performance gap between each of the input modalities. The ECM models obtained between 40-50\% of the mAP of the RGB frames across all of the models. The AP 50 fared slightly better, demonstrating that the model had more difficulty producing precise bounding boxes when trained on the ECMs. Additionally, the RVT model performed worse than any of the other detectors, despite being specifically constructed for processing event streams. This model even includes a temporal component, allowing it to access information across frames. This suggests that the benefits of large scale pre-training outweigh the benefits of event-specific architectures, despite the large distribution shift between RGB frames and ECMs.

\begin{table}[]
    \centering
    \begin{tabular}{l l l}
    \toprule
        Input & Pre-training Type & mAP \\\midrule
        \multirow{3}{*}{RGB Frame} & Backbone + Full Detector & 0.128 \\
        & Backbone & 0.120 \\
        & None & 0.083 \\\hline
        \multirow{3}{*}{ECM} & Backbone + Full Detector & 0.063 \\
        & Backbone & 0.057 \\
        & None & 0.047 \\
        \bottomrule
    \end{tabular}
    \caption{VidDrone-VID dev test set performance using Cascade RCNN under various pre-training regimes.}
    \label{tab:pretraining}
\end{table}
While there are likely many reasons for this large performance gap, we hypothesized that part of it was due to a lack of alignment between the pre-training data used for these detectors, and our ECMs. All of the models used an ImageNet \cite{imagenet} backbone and were fully pre-trained on MSCOCO \cite{mscoco}. We conducted two additional experiments to verify this hypothesis. In the first, we explored removing both the MSCOCO pre-training and the ImageNet pre-training for each of the modalities using the Cascade RCNN model; the results can be found in Table \ref{tab:pretraining}. We found that removing all pre-training from the RGB frame detector decreased performance by 35.2\%, while removing it for the ECM backbone resulted in a 25.4\% decrease. The large performance drop for RGB frames suggests that the Cascade RCNN detector was more reliant upon the pre-trained weights when trained with RGB frames compared to the ECM input, further suggesting that the RGB frames are better aligned with the pre-training data. Additionally, we explored the impact of pre-training on the RVT model, which did not have any pre-training. We pre-trained the model using the Gen4 dataset \cite{gen4}, then fine-tuned the model on the VisDrone event streams and found that mAP improved from 0.046 to 0.054. This suggests that pre-training on event-specific datasets can improve downstream performance. However, this performance is still worse than using ImageNet + MSCOCO pre-trained weights.

\subsection{Reconstructed Frames vs. RGB Frame Comparison}
\begin{table}[]
    \centering
    \begin{tabular}{l l l l}
    \toprule
        Detector & Event Rep. & mAP & AP 50 \\\midrule
        \multirow{4}{2em}{Cascade RCNN} & ECM & 0.063 & 0.132 \\
        & FireNet & 0.095 & 0.198 \\
        & FlowNet & 0.098 & 0.207 \\
        & FlowNet-FT & 0.107 & 0.218 \\\hline
        \multirow{4}{*}{DINO} & ECM & 0.071 & 0.155 \\
        & FireNet & 0.122 & 0.265 \\
        & FlowNet & 0.126 & 0.276 \\
        & FlowNet-FT & 0.137 & 0.291\\
        \bottomrule
    \end{tabular}
    \caption{Event stream detection results comparing ECM inputs to FireNet and FlowNet inputs (FT=Fine-tuned).}
    \label{tab:firenet}
\end{table}
Rather than pre-training on a large event stream dataset, we propose better aligning the event representations with ImageNet/MSCOCO pre-training data. While the ECMs are easily interpreted by humans, albeit not as effectively as RGB frames, the pixel values, which are sparse and based upon brightness, differ significantly from standard images. Therefore, as discussed in Section \ref{sec:method}, we employ FireNet and FlowNet to convert our event streams into gray-scale images, better aligning them with standard images.

The results of these experiments can be seen in Table \ref{tab:firenet}. Processing the event stream using FireNet and FlowNet pre-trained weights leads to significant improvements in performance, with further improvement when we fine-tune on our VisDrone event data (FlowNet-FT); results with Cascade R-CNN improved by 69.8\% and with DINO by 93.0\%. Although these models do not outperform the RGB frames, they do close the gap in performance, where models trained on the reconstructed event frames achieved approximately 80\% of the RGB performance. Furthermore, the FlowNet + DINO model exceeds the performance of the Cascade RCNN that uses RGB frames. Cascade RCNN is a strong baseline detector; this result suggests that utilizing an event-to-frame conversion tool along with a state-of-the-art detector can even achieve better results than many frequently used detectors on RGB frames in our overhead setting.

\section{Conclusion}
In this work, we investigated the viability of event streams for overhead object detection. We found that there was a large performance gap between detectors trained on RGB frames and detectors trained on dense event representations. We demonstrated that this performance gap was due to a greater disparity between the event representations and the pre-training data used for our object detectors. We then proposed using FireNet and FlowNet to convert the event streams to gray-scale video and presented experiments demonstrating that this significantly improved mAP. These results suggest that although architectural improvements specific to event streams are one viable method for improving object detection performance, an alternative is better aligning event representations with standard image representations to take advantage of powerful pre-trained deep learning models.

\section{Acknowledgements}
This material is based upon work supported by the Department of Energy, Office of Science, Advanced Scientific Computing Research program, under award number 77902 and partially funded by the PNNL Cloud, HPC, and Edge for Science and Security (CHESS) Lab-Directed Research \& Development initiative.

\bibliographystyle{IEEEbib}
\bibliography{ssiai}

\end{document}